%% file: TMM/main.tex
\documentclass[journal,10pt]{IEEEtran}

\usepackage{amsmath,amsfonts}
\usepackage{amssymb}
\usepackage{algorithmic}
\usepackage{algorithm}
\usepackage{array}
\usepackage{xcolor}
\usepackage{graphicx}
\usepackage{multirow}
\usepackage{colortbl}
\usepackage{booktabs} 
\usepackage{url}
\usepackage{cite}
\usepackage[breaklinks=true,colorlinks,citecolor=blue]{hyperref}

\graphicspath{{./Imgs/}}
\DeclareGraphicsExtensions{.pdf,.jpg,.png}

\AtEndPreamble{
    \usepackage[capitalize]{cleveref}
    \crefname{section}{Sec.}{Secs.}
    \Crefname{section}{Section}{Sections}
    \crefname{figure}{Fig.}{Figs.}
    \Crefname{figure}{Figure}{Figures}
    \crefname{table}{Tab.}{Tabs.}
    \Crefname{table}{Table}{Tables}
}

\def\ie{\emph{i.e.}}
\def\eg{\emph{e.g.}}

\newcommand{\para}[1]{\vspace{0.05in}\noindent{\bf #1}\quad}

\begin{document}
	
\title{Make It Up: Fake Images, Real Gains in Generalized Few-shot Semantic Segmentation}

% \author{
    %   \IEEEauthorblockN{Author Name 1\IEEEauthorrefmark{1}, Author Name 2\IEEEauthorrefmark{2}, and Author Name 3\IEEEauthorrefmark{3}}
    %   \IEEEauthorblockA{\IEEEauthorrefmark{1}Department of Electrical Engineering, University Name, City, Country \\
        %     Email: author1@university.edu}
    %   \IEEEauthorblockA{\IEEEauthorrefmark{2}Department of Computer Science, University Name, City, Country \\
        %     Email: author2@university.edu}
    %   \IEEEauthorblockA{\IEEEauthorrefmark{3}Department of Mechanical Engineering, University Name, City, Country \\
        %     Email: author3@university.edu}
    %   \IEEEauthorblockN{IEEE Publication Technology,~\IEEEmembership{Staff,~IEEE,}}
    % }

\author{Guohuan Xie, Xin He, Dingying Fan, Le Zhang, Ming-Ming Cheng,~\IEEEmembership{Senior Member, IEEE} and Yun Liu
    % <-this % stops a space
    \thanks{This work is supported in part by the Fundamental Research Funds for the Central Universities (Nankai University, No. 070-63253235) and in part by NSFC (No. 62576176). The computational resources are supported by the Supercomputing Center of Nankai University (NKSC). \textit{(Corresponding author: Yun Liu)}}%
    \thanks{G. Xie, and D. Fan are with the College of Software, Nankai University, Tianjin 300350, China. (e-mail: xieguohuan@mail.nankai.edu.cn; dingyingfan@mail.nankai.edu.cn)}%
    \thanks{X. He is with the School of Computer Science and Engineering, Tianjin University of Technology, Tianjin 300384, China. (e-mail: xhe@email.tjut.edu.cn)}%
    \thanks{L. Zhang is with School of Information and Communication Engineering, UESTC, Chengdu 611731, China. (e-mail: lezhang@uestc.edu.cn)}%
    \thanks{M.M. Cheng and Y. Liu are with VCIP, College of Computer Science, and Academy for Advanced Interdisciplinary Studies, Nankai University, Tianjin 300350, China, and also with NKIARI, Shenzhen Futian, Shenzhen 518045, China. (e-mail: cmm@nankai.edu.cn; liuyun@nankai.edu.cn)}% <-this % stops a space.
}



\maketitle

\begin{abstract}
Generalized few-shot semantic segmentation (GFSS) is fundamentally limited by the coverage of novel-class appearances under scarce annotations. While diffusion models can synthesize novel-class images at scale, practical gains are often hindered by insufficient coverage and noisy supervision when masks are unavailable or unreliable. We propose Syn4Seg, a generation-enhanced GFSS framework designed to expand novel-class coverage while improving pseudo-label quality. Syn4Seg first maximizes prompt-space coverage by constructing an embedding-deduplicated prompt bank for each novel class, yielding diverse yet class-consistent synthetic images. It then performs support-guided pseudo-label estimation via a two-stage refinement that i) filters low-consistency regions to obtain high-precision seeds and ii) relabels uncertain pixels with image-adaptive prototypes that combine global (support) and local (image) statistics. Finally, we refine only boundary-band and unlabeled pixels using a constrained SAM-based update to improve contour fidelity without overwriting high-confidence interiors. Extensive experiments on PASCAL-$5^i$ and COCO-$20^i$ demonstrate consistent improvements in both 1-shot and 5-shot settings, highlighting synthetic data as a scalable path for GFSS with reliable masks and precise boundaries.
\end{abstract}

\input{sec/1_intro}
\input{sec/2_relatedwork}
\input{sec/3_method}

\input{sec/4_experiments}

\input{sec/5_conclu}

% \section*{Acknowledgment}
% This work is supported in part by the Fundamental Research Funds for the Central Universities (Nankai University, No. 070-63253235) and in part by NSFC (No. 62576176). The computational resources are supported by the Supercomputing Center of Nankai University (NKSC).

{\small
\bibliographystyle{IEEEtran}
\bibliography{main}
}

\vspace{-.2in}

\begin{IEEEbiography}
[{\includegraphics[width=1in,height=1.25in,clip,keepaspectratio]{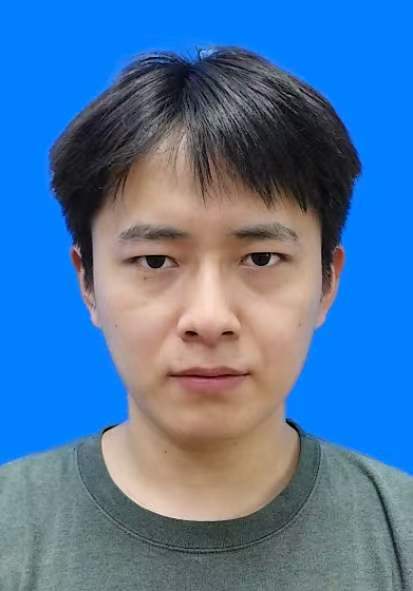}}]
{Guohuan Xie} is currently pursuing his B.E. degree at the College of Software, Nankai University, from 2022 to 2026. He is working with Prof. Yun Liu. His research interests include computer vision and machine learning.
\end{IEEEbiography}

\vspace{-.2in}

\begin{IEEEbiography}
[{\includegraphics[width=1in,height=1.25in,clip,keepaspectratio]{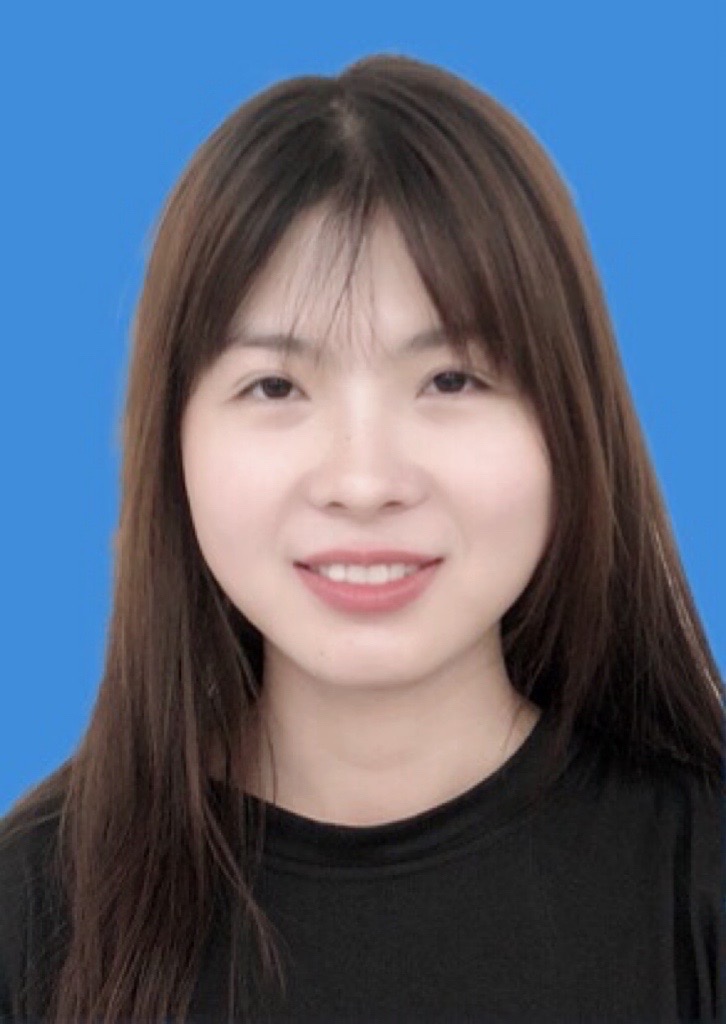}}]
{Xin He} received her B.E. and Ph.D. degrees from the School of Computer Science and Technology, China University of Mining and Technology in 2019 and 2025, respectively. Currently, she is a lecturer at the School of Computer Science and Engineering, Tianjin University of Technology. Her research interests include computer vision and semantic segmentation for remote sensing images.
\end{IEEEbiography}

\vspace{-.2in}

\begin{IEEEbiography}
[{\includegraphics[width=1in,height=1.25in,clip,keepaspectratio]{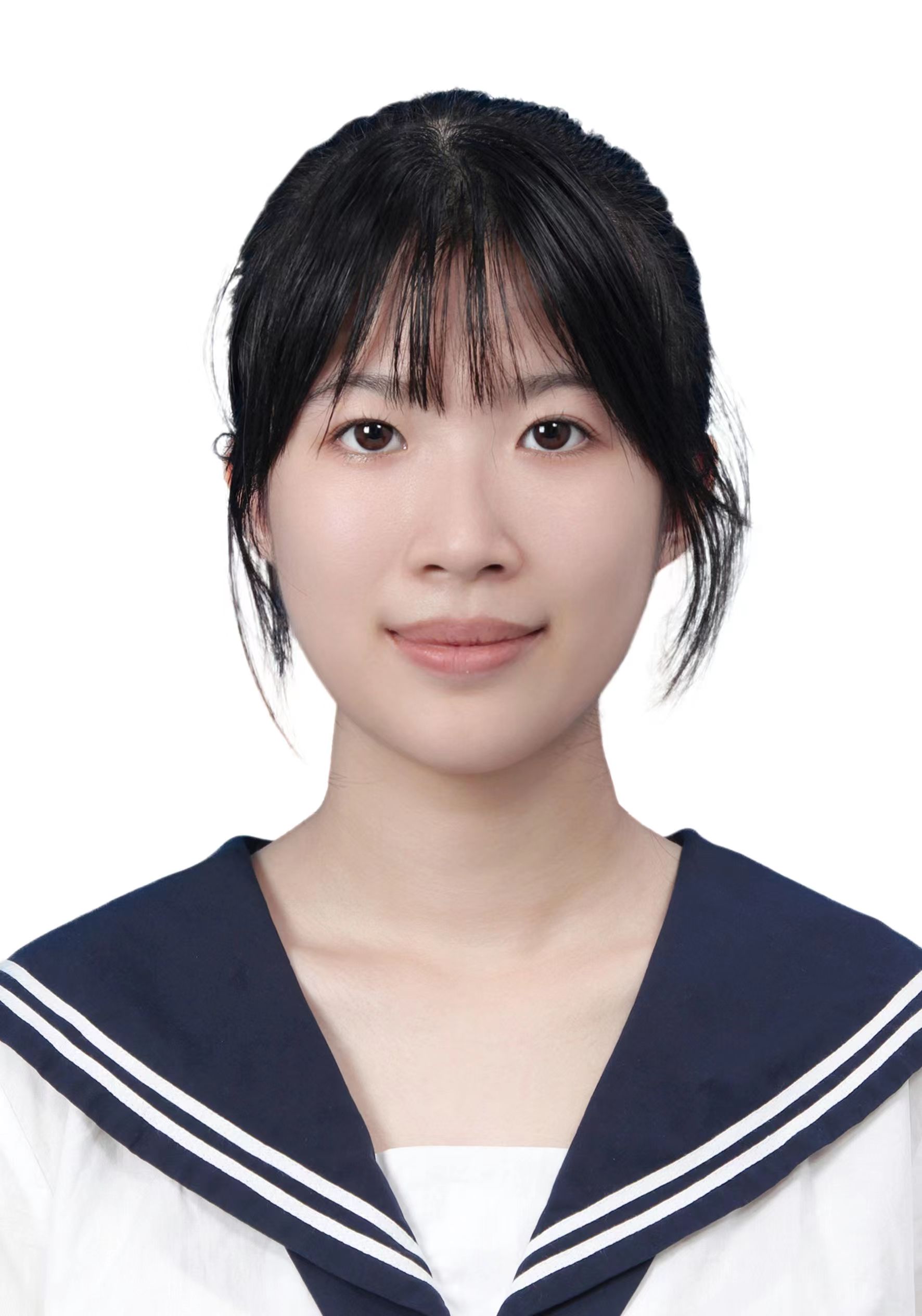}}]
{Dingying Fan} is currently pursuing the B.E. degree at the College of Software, Nankai University, Tianjin, China, where she is working with Prof. Yun Liu. Her research interests include computer vision, large language models, and multi-agent systems.
\end{IEEEbiography}

% \vspace{-.2in}

\begin{IEEEbiography}[{\includegraphics[width=1in,height=1.25in,clip,keepaspectratio]{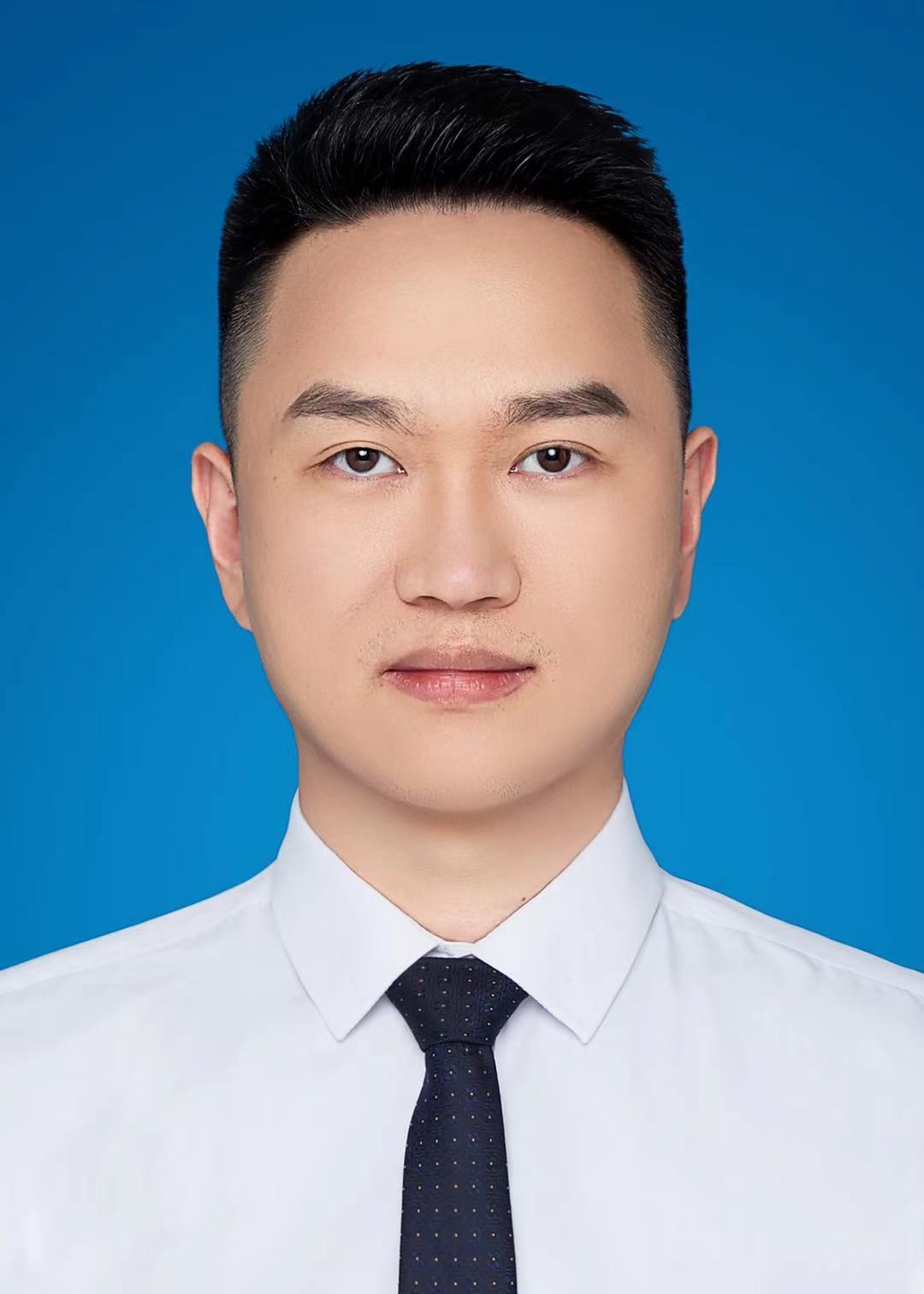}}]{Le Zhang} is a professor with the School of Information and Communication Engineering at the University of Electronic Science and Technology of China (UESTC). He earned his Ph.D. degree from Nanyang Technological University (NTU) in 2016. Following his graduation, he served as a postdoctoral fellow at the Advanced Digital Sciences Center (ADSC) in Singapore from 2016 to 2018. His career continued to evolve as he took on the role of a research scientist at the Institute for Infocomm Research (I2R) under the Agency for Science, Technology, and Research (A*STAR), Singapore, where he worked from 2018 to 2021. His research primarily focuses on computer vision and machine learning. He is an associate editor of Neural Networks, Neurocomputing, and IET Biometrics. Additionally, he has been a guest editor for several journals, including IEEE Transactions on Neural Networks and Learning Systems, IEEE Transactions on Big Data, Pattern Recognition and so on. He has been honored with multiple paper awards, including the 2022 Norbert Wiener Review Award in IEEE/CAA J. Autom. Sinica, as well as the Best Paper Awards at the 2022 IEEE HPCC and IEEE ICIEA conferences.
\end{IEEEbiography}

\vspace{-.2in}

\begin{IEEEbiography}
[{\includegraphics[width=1in,height=1.25in,clip,keepaspectratio]{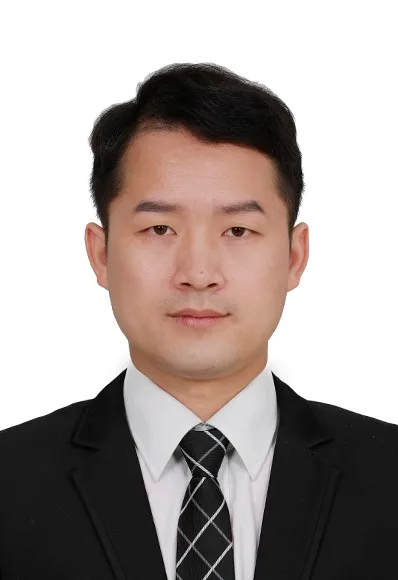}}]
{Ming-Ming Cheng} (Senior Member, IEEE) received the PhD degree from Tsinghua University, in 2012, and then worked with Prof. Philip Torr in Oxford for 2 years. Since 2016, he is a full professor at Nankai University, leading the Media Computing Lab. His research interests include computer vision and computer graphics. He received research awards, including ACM China Rising Star Award,
IBM Global SUR Award, and CCF-Intel Young Faculty Researcher Program. He is a senior member of the IEEE and on the editorial boards of IEEE Transactions on Pattern Analysis and Machine Intelligence and IEEE Transactions on Image Processing.
\end{IEEEbiography}

\vspace{-.2in}

\begin{IEEEbiography}
[{\includegraphics[width=1in,height=1.25in,clip,keepaspectratio]{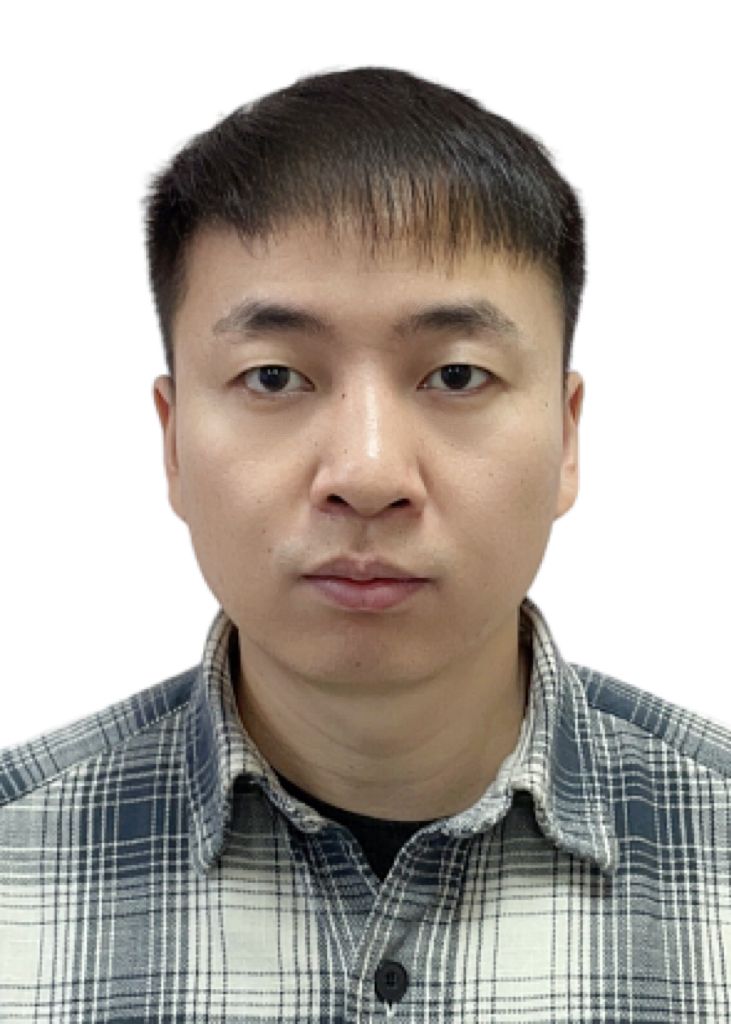}}]
{Yun Liu} received his B.E. and Ph.D. degrees from Nankai University in 2016 and 2020, respectively. Then, he worked with Prof. Luc Van Gool as a postdoctoral scholar at the Computer Vision Lab, ETH Zurich, Switzerland. After that, he worked as a senior scientist at the Institute for Infocomm Research (I2R), A*STAR, Singapore. Currently, he is a professor at the College of Computer Science, Nankai University. His research interests include computer vision and deep learning.
\end{IEEEbiography}

\vfill

\end{document}

%% file: sec/1_intro.tex
\section{Introduction}
\begin{figure}[t]
    \centering
    \includegraphics[width=\columnwidth]{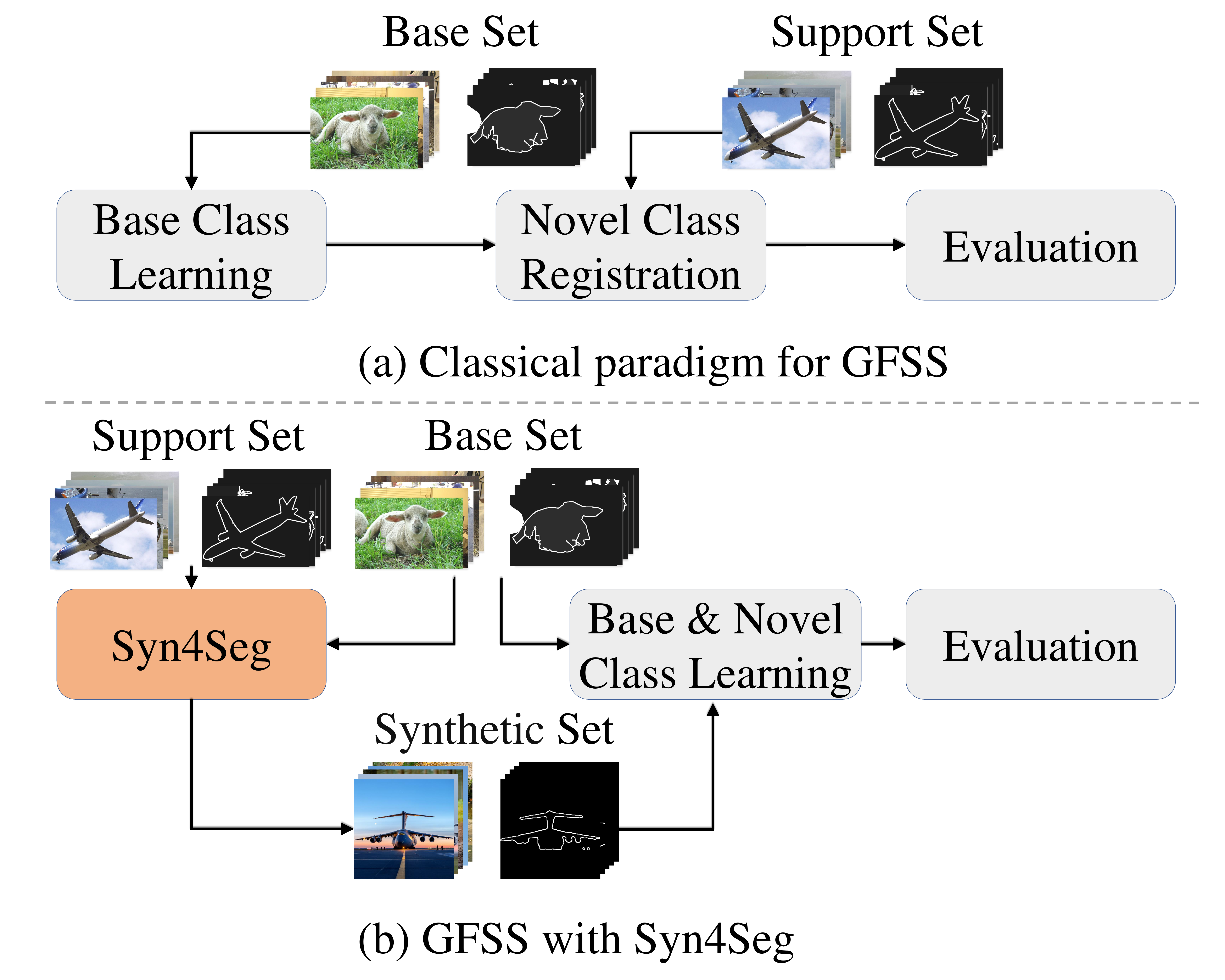}
    \caption{Pipelines of classical GFSS works and our Syn4Seg framework.
(a) Classical GFSS methods follow three stages: base class learning, novel class registration, and evaluation, where novel classes rely on limited manual annotations, resulting in low diversity and constrained performance.
(b) Our Syn4Seg framework synthesizes diverse, realistic novel class images and pseudo labels guided by support annotations. These synthesized samples, together with base data, are used to train a semantic segmentation model, enabling robust recognition of both base and novel classes during evaluation.}
    \label{fig:comparision}
\end{figure}

Few-shot semantic segmentation (FSS) enables models to segment unseen classes with only a few annotated support examples, reflecting real-world scenarios where pixel-level annotations are expensive and scarce. 
However, conventional FSS frameworks depend on class-specific support sets: each novel class requires a dedicated support set and separate inference passes \cite{boudiaf2021few,lu2021simpler,lu2023prediction,yang2021free,alayrac2022flamingo,peng2025sam}, which severely limits scalability and efficiency. 
To address these issues, Generalized Few-shot Semantic Segmentation (GFSS) \cite{tian2022generalized} has been proposed to jointly segment both base and novel classes within a single inference pass. 
Despite its effectiveness, GFSS remains fundamentally constrained by the limited diversity of manually annotated support data. 
While base classes are typically well-annotated and diverse, novel classes are learned from only a handful of annotated support examples, as shown in \cref{fig:comparision}(a), resulting in limited intra-class variation and poor generalization to unseen instances.

Recent advances in diffusion-based image generation \cite{nichol2021glide,rombach2022high,tian2024visual,yin2024improved} have enabled the synthesis of images that closely approximate real-world distributions. 
This technological progress offers a promising way to alleviate the diversity bottleneck in GFSS by enriching novel class representations beyond the constraints of limited annotations. 
However, effectively leveraging synthetic images for GFSS introduces three major challenges: 
(1) Directly generating synthetic images based on class names or simple prompts containing class labels often leads to redundancy in the generated images, resulting in a lack of diversity. This insufficient diversity limits the model's exposure to novel visual patterns, making it crucial to generate sufficiently varied novel class samples that better capture real-world variations.
(2) Current methods struggle to directly generate accurate segmentation masks for synthetic images. Using predicted masks from a pre-trained GFSS model inevitably results in low-quality, poorly aligned masks~\cite{zhang2019decoupled,zhou2020sal,liu2021cross,xu2023wave}. These issues introduce significant label noise, severely degrading segmentation performance, and highlighting the need for high-quality, class-consistent segmentation masks in the synthesis process.
(3) Additionally, segmentation masks generated through semantic segmentation models often suffer from poor boundary quality \cite{yu2018learning, wang2022active, li2020improving, takikawa2019gated, zhen2020joint,yin2022contour,zhou2024boundary}. This imprecision in mask boundaries causes pixel misalignments along object boundaries, which in turn lowers the Intersection over Union (IoU), increases false positives and negatives, and disrupts the learning of fine-grained spatial cues essential for accurate segmentation.

To address these challenges, we propose \textbf{Syn4Seg}, a generation-enhanced framework for GFSS, as shown in \cref{fig:comparision}(b).
Specifically, we introduce \textbf{High-quality Diverse Image Generation (HDIG)} by conditioning diffusion-based generation on category prototypes, which can mitigate mode collapse and redundancy, enriching both the visual and semantic coverage of novel categories and exposing the segmentation model to a broader range of variations. After generating these novel class images by HDIG, a straightforward approach is to obtain their semantic segmentation masks directly using a pre-trained GFSS model. However, such pseudo-masks often suffer from semantic noise and misaligned regions, which can degrade training if used naively. To address this issue, we propose \textbf{Adaptive Pseudo-label Enhancement (APE)}, which refines pseudo-masks through two complementary processes. \textit{Adaptive Pseudo-label Filtering (APF)} removes low-confidence regions guided by prototype similarity, while \textit{Adaptive Pseudo-label Relabeling (APR)} corrects inconsistent areas based on contextual cues. This dual-stage adaptation significantly improves the semantic consistency and reliability of pseudo-masks, providing higher-quality supervision for downstream segmentation. But even after enhancement by APE, pseudo-masks typically exhibit coarse or blurry boundaries, which hinder accurate object delineation. These imprecise boundaries can introduce pixel-level labeling errors, particularly along object contours, leading to decreased IoU, more false positives and negatives, and impaired learning of fine-grained spatial cues that are critical for precise segmentation \cite{li2020improving,takikawa2019gated,zhen2020joint}. To address this,  we propose \textbf{SAM-based Boundary Refinement (SABR)}, which can leverage the strong localization and boundary-awareness of the Segment Anything Model (SAM) \cite{kirillov2023segment} to refine uncertain boundaries, producing sharp and spatially coherent contours that better align with object structures.

Extensive experiments demonstrate that Syn4Seg consistently outperforms existing state-of-the-art methods across various settings. Notably, under the challenging 5-shot condition, Syn4Seg achieves improvements of 2.20\% and 3.69\% in mean mIoU, and 3.28\% and 4.34\% in hIoU, on the PASCAL-$5^i$ and COCO-$20^i$ datasets than state-of-the-art methods \cite{Sakai2024ASS,hossain2024visual}. These results clearly validate the effectiveness of our generation-enhanced approach.

Our main contributions can be summarized as follows:
\begin{itemize}
\item We propose \textbf{Syn4Seg}, a simple yet effective generation-enhanced GFSS framework that synthesizes class-aligned, diverse samples via \textbf{HDIG} to mitigate data scarcity and improve novel class generalization.
\item We introduce the \textbf{APE} mechanism that jointly performs prototype-guided APF and APR, enabling adaptive enhancement of pseudo-masks with support-aware consistency and improved semantic reliability.
\item We develop the \textbf{SABR} module to accurately delineate object contours in uncertain regions, achieving high-quality, well-aligned, and visually consistent pseudo-masks that further boost segmentation performance.
\end{itemize}

%% file: sec/2_relatedwork.tex
\section{Related work}

\noindent\textbf{Few-shot Semantic Segmentation.}\quad
Few-shot semantic segmentation methods fall into two main branches: prototype-based and dense comparison-based. Prototype-based approaches \cite{liu2020part,yang2020prototype,hossain2024visual,zhang2022feature} construct compact class representations (prototypes) from a few support examples and fuse them with query features via attention mechanisms or iterative refinement. This design provides stable, class-level guidance and strong generalization for recognizing target regions in the query image. In contrast, dense comparison-based approaches \cite{zhang2019pyramid,min2021hypercorrelation,wang2020few,boudiaf2020information,zhu2023transductive} perform pixel- or region-level matching between support and query images. By preserving spatial detail and local correspondences, these methods handle large intra-class variations better. However, they incur higher computational costs and are more sensitive to feature misalignment. Despite their progress, conventional FSS frameworks are limited in that they are designed to segment only one novel class at a time \cite{zhang2018sg,qiao2019transductive,li2020fss,fan2022self,liu2022learning,tan2023diffss,zhu2024unleashing,liu2024bidirectional,tong25b,xu25an}. Each support set corresponds to a single target class, and the model cannot simultaneously recognize both base and novel classes within the same query image. As a result, segmenting all potential novel classes requires multiple inference passes with different support sets, limiting their scalability and practicality in open or mixed-category scenarios.

\para{Generalized Few-shot Semantic Segmentation.}
Generalized few-shot semantic segmentation (GFSS) extends FSS by requiring a model to adapt to novel classes from only $k$-shot exemplars while segmenting the full label space, including both base and novel classes, at test time \cite{tian2022generalized}. Compared with standard FSS, this setting is more practical but also more challenging, since the model must absorb novel-class knowledge from extremely limited supervision while preserving strong discrimination over well-learned base classes. Existing GFSS methods often address this problem by introducing learned prompts, prototype adaptation, or lightweight adapters to inject novel-class cues into pretrained segmentation decoders \cite{hajimiri2023strong,hossain2024visual,Liu2023LearningOP,Sakai2024ASS,liu2024harmonizing}. While effective, these methods still rely on scarce manually annotated support samples, limiting the diversity and coverage of novel-class supervision. Recent work \cite{qiu2024aligndiff} has explored diffusion-based synthesis to augment GFSS with generated novel-class samples, opening a promising direction for alleviating data scarcity. However, directly leveraging generative models in GFSS remains challenging: images generated from simple class-name prompts often lack sufficient diversity, limiting coverage of intra-class variations, while the corresponding masks obtained from pretrained segmentation models may contain semantic noise, misaligned regions, or inaccurate boundaries. These issues restrict the effectiveness of synthetic supervision and weaken the benefits of generative augmentation for improving generalization and segmentation quality. Motivated by these limitations, we propose a generative synthesis framework that couples diverse diffusion-based image generation with principled mask refinement, enabling the construction of high-quality, semantically aligned image--mask pairs for novel classes and thereby improving novel-class generalization while preserving strong base-class performance.

% \para{Promptable Segmentation.}
% Promptable segmentation frames mask prediction as a prompt-conditioned generation task, where a single model produces target regions based on various spatial or semantic cues (\eg, points, boxes, or text). Large-scale prompt-driven pre-training and flexible decoders, as in SAM \cite{kirillov2023segment}, have substantially advanced zero-shot and interactive segmentation. Further developments, such as multimodal, memory-aware decoders (\eg, SEEM \cite{zou2023segment}) and text- or detector-guided approaches (\eg, Grounded SAM \cite{ren2024grounded}, CLIPSeg \cite{luddecke2022image}, LLaFS++ \cite{zhu2025llafs++}), demonstrate how language or detector outputs can serve as open-vocabulary prompts.
% Building on these foundations, our SAM-based Boundary Refinement module explores a complementary direction: rather than introducing new prompt types or modalities, it leverages SAM as a high-quality refinement engine to enhance coarse or noisy masks. By extracting prototype-guided candidate boxes and refining them with SAM, our method produces large-scale, high-quality synthetic annotations that strengthen novel-class generalization while maintaining base-class performance.

%% file: sec/3_method.tex
\section{Methodology}
\begin{figure*}[t]
    \centering
    \includegraphics[width=\textwidth]{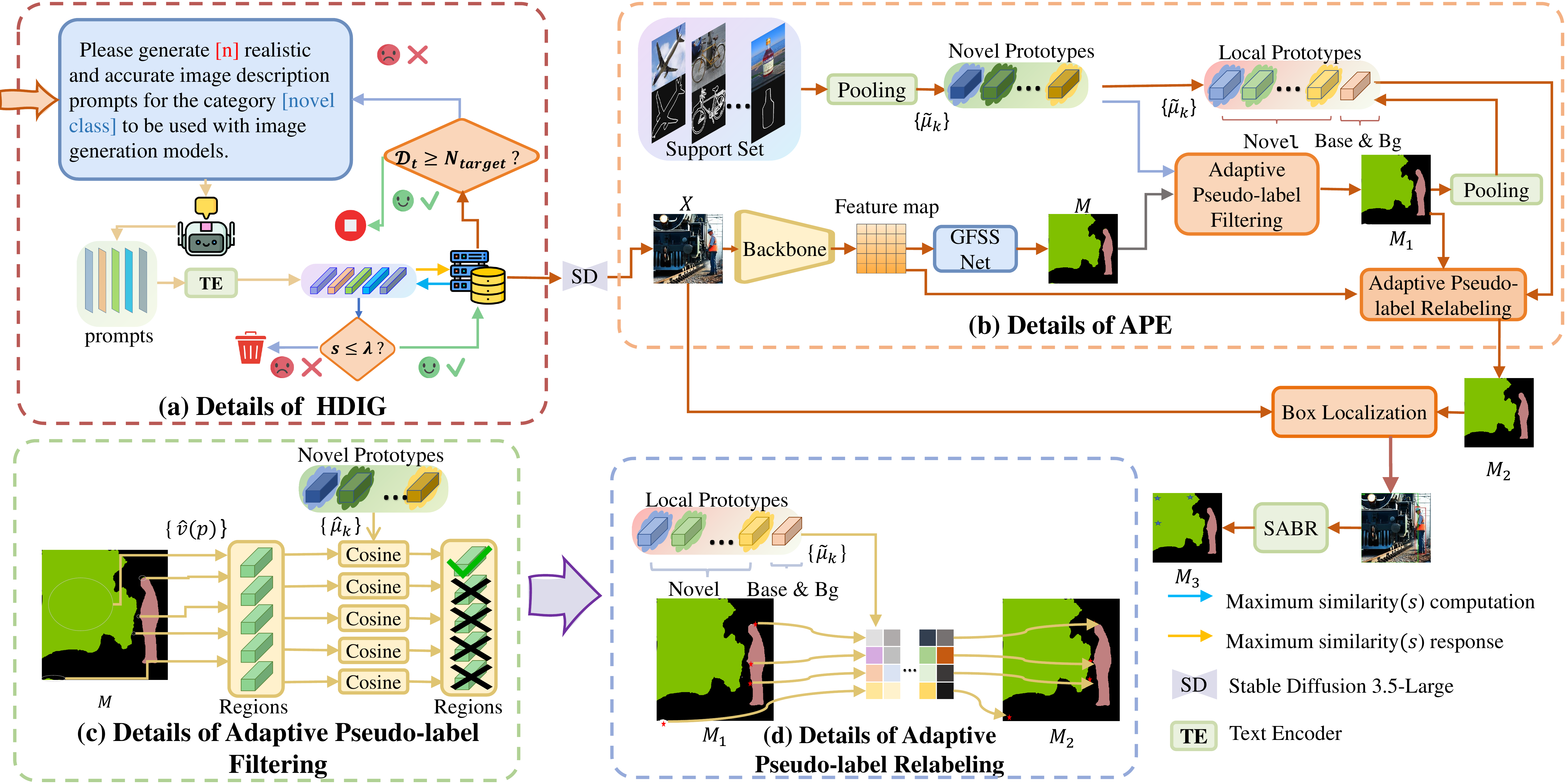}
    \caption{(a), (b) Given several novel classes and their support set, we first use HDIG to synthesize high-quality, diverse novel class images proportional to the base class images. Guided by support images, APE refines these synthesized images to generate accurate masks through two stages: Adaptive Pseudo-label Filtering (APF) and Adaptive Pseudo-label Relabeling (APR). Specifically, APF filters noisy regions in the initial mask \(M\) to obtain \(M_1\); APR relabels these unlabeled regions to yield \(M_2\). Finally, SAM-based Boundary Refinement (SABR) further refines the mask boundaries, yielding the final \(M_3\). (c), (d) illustrate the details of APF and APR.}
    \label{fig:overview}
    \vspace{-1em}
\end{figure*}

We propose the Syn4Seg framework as shown in \cref{fig:overview}, which alleviates the shortage of novel class images for GFSS scenarios in real-world applications by synthesizing a large number of high-quality and diverse novel class images through HDIG. For these synthetic images, we then employ APE and SABR to produce their high-quality masks. APE comprises two stages: Adaptive Pseudo-label Filtering and Adaptive Pseudo-label Relabeling. Through APE and SABR, we obtain abundant novel class images together with their corresponding high-quality masks. Ablation experiments in \cref{sec:ablation} demonstrate the effectiveness of HDIG, APE, and SABR.

\subsection{Problem Formulation}
GFSS considers a label space partitioned into disjoint base and novel sets \(C_b\) and \(C_n\) with \(C_b \cap C_n=\varnothing\). During development, we are given a large labeled base dataset $\mathcal{D}_b$ for classes in $C_b$, enabling the model to learn to segment the base classes. For each class $c \in C_n$, we are provided with a small $K$-shot support set $\mathbf{S}_c = \{(\mathbf{S}_i^c, \mathbf{M}_i^c)\}_{i=1}^K$ of image--mask pairs. In total, there are $|C_n|$ such $K$-shot support sets corresponding to the novel classes. The model is adapted using these $|C_n|$ support sets to segment the novel classes.
 The objective of GFSS is to train a single segmentation model that (i) retains strong performance on base classes and (ii) incorporates the limited \(K\)-shot information for novel classes. This enables the model to predict pixel-wise labels from the union \(C_b\cup C_n\) for every query image without per-image support at test time.

\subsection{High-quality Diverse Image Generation}
\label{sec:Syn4Seg_overview}
Directly using class names or simple textual prompts for image synthesis often leads to limited diversity and high visual similarity among synthetic samples. To address this limitation and improve the representation of novel classes in GFSS, we aim to generate synthetic images that are both class-consistent and visually diverse. 
Given that the quality and diversity of generated images are largely determined by the textual prompts used to guide the diffusion model, the first step of our framework is to construct, for each novel class, a prompt set that remains centered on the target class while covering semantically diverse descriptions.

To construct a prompt collection that is both centered on a target novel class \(c\) and semantically diverse, \textbf{HDIG} adopts an \textit{iterative prompt generation} strategy. At the \(t\)-th prompt generation round, the agent generates \(s\) candidate prompts $\mathcal{P}_t=\{p_{t,1},p_{t,2},\dots,p_{t,s}\}$, where the agent architecture is shown in \cref{fig:overview}(a), and the underlying base model used to implement the agent is DeepSeek-V3.
Our goal is to ensure diversity between different prompts. This naturally leads us to consider the cosine similarity between different prompt embeddings, filtering out those with high similarity and retaining those with lower similarity. To ensure that the prompt embeddings are consistent with the semantic space of Stable Diffusion 3.5-Large (SD3.5), the model we use for generating images, we leverage the text encoder of SD3.5, which guarantees semantic alignment across the embeddings. Specifically, each prompt is encoded using the dual-encoder text pipeline of SD3.5 (CLIP-L and CLIP-G). The outputs are pooled and concatenated into a 2048-dimensional representation, consistent with the architecture of SD3.5. We use the \emph{normalized} text embedding \(\tilde{\varphi}(p)\in\mathbb{R}^d\) for similarity evaluation, where \(\tilde{\varphi}(p)\) is obtained by applying \(\ell_2\)-normalization to the encoder output. Let \(\mathcal{D}_t^{(i-1)}\) represent the progressively updated database within the batch, initialized as \(\mathcal{D}_t^{(0)}=\mathcal{D}_{t-1}\). The maximum cosine similarity between the current candidate and existing entries is computed as:
\begin{equation}
s_{t,i}=\max_{u\in\mathcal{D}_t^{(i-1)}} \tilde{\varphi}(p_{t,i})^\top u,\qquad i=1,\dots,s,
\end{equation}
where \(u\) denotes a stored normalized embedding in \(\mathcal{D}_t^{(i-1)}\).

\begin{figure}[t]
    \centering
    \includegraphics[width=\columnwidth]{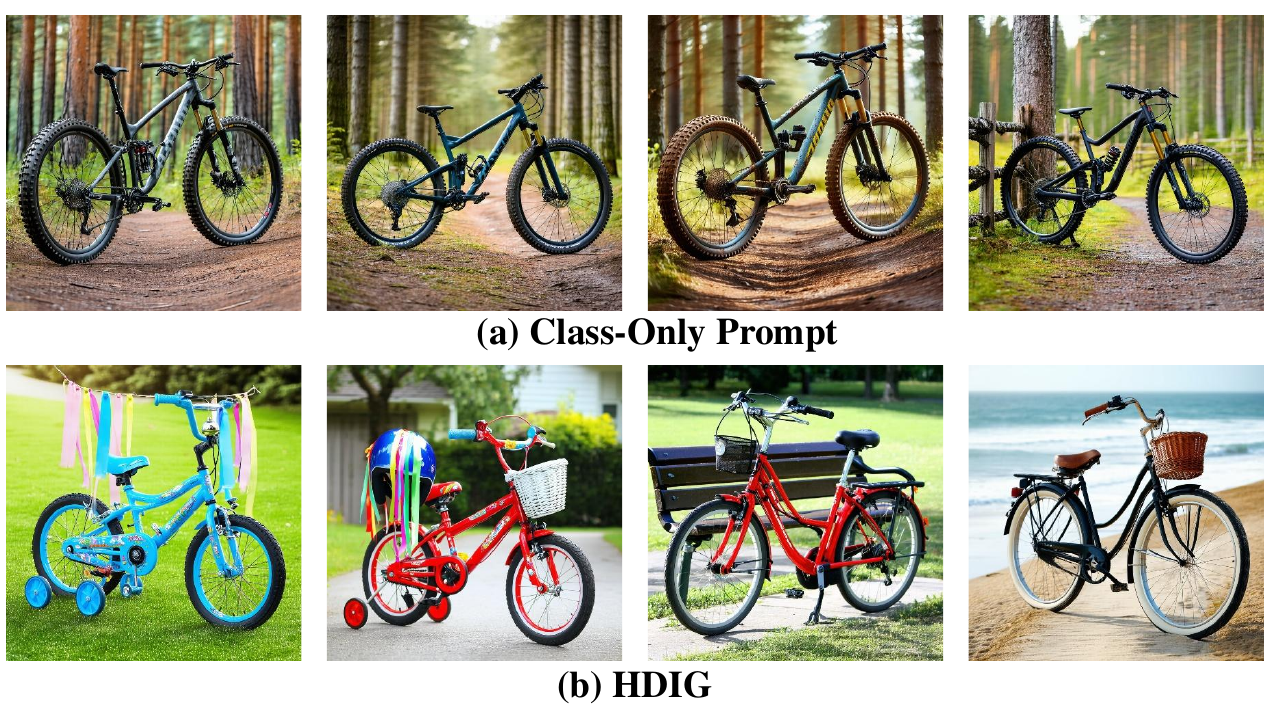}
    \caption{Qualitative comparison between images generated directly using class-name prompts (a) and those generated with our HDIG method (b).}
    \label{fig:ourhdig}
\end{figure}

Given a diversity threshold $\lambda\in[-1,1]$, a candidate prompt is accepted
only if its similarity score $s_{t,i}$ is below $\lambda$, ensuring sufficient
diversity with respect to the existing database. Accepted prompts are immediately
inserted into the database, and after processing the entire batch we obtain
$\mathcal{D}_t=\mathcal{D}_t^{(s)}$. The generation process terminates once the database reaches a target size $N_{\text{target}}$. We set this target to approximately match the average number of images per class in the original training set, \ie, \textbf{300 per
class} for PASCAL-$5^i$ and \textbf{1000 per class} for COCO-$20^i$. All accepted prompts are then used to synthesize images at a
resolution of $768\times768$ using SD3.5, which are subsequently processed by
the GFSS network to produce the initial semantic masks. A qualitative comparison between images generated directly using class-name prompts and those generated with our HDIG method is provided in the Appendix, where HDIG exhibits substantially improved visual diversity while preserving class consistency. A qualitative comparison between images generated directly using class-name prompts and those generated with our HDIG method is shown in \cref{fig:ourhdig}. Compared with class-only prompts, HDIG produces images with substantially richer visual diversity while preserving class consistency, which supports its effectiveness for enriching novel-class coverage.

\subsection{Adaptive Pseudo-label Enhancement}
\label{sec:pas}

A straightforward way to use the generated novel-class images and the initial semantic masks from the GFSS network is to proportionally mix them with the original dataset and then train the downstream semantic segmentation network. However, these initial masks are noisy, which hinders learning robust novel class representations for the model. Moreover, noisy pseudo-labels can propagate GFSS network errors into the downstream semantic segmentation network. Motivated by \cite{an2025generalized}, we propose an effective pseudo-mask enhancement paradigm, APE, which leverages a small set of valuable support samples to further refine the initial masks and thereby significantly improve mask quality. APE consists of \textit{Adaptive Pseudo-label Filtering (APF)} and \textit{Adaptive Pseudo-label Relabeling (APR)}, which we detail below.

\subsubsection{Adaptive Pseudo-label Filtering}
\label{sec:APF}
APF employs a compact support set to discard unreliable pseudo-labels and improve the overall quality of novel class masks. Let the support set be \(\mathcal{S}=\{(\mathbf{X_j},\mathbf{M_j})\}_{j=1}^{N_S}\), where \(\mathbf{X_j}\) is a support image and \(\mathbf{M_j}\) corresponds  to its pixel-wise mask. For each support image, we extract its feature map \(F(\mathbf{X_j})\in\mathbb{R}^{C\times H\times W}\) from the backbone for each support image, and compute the prototype for novel class \(k\) by averaging features over pixels labeled as \(k\):
\begin{equation}
\mathbf{\mu_k} = \frac{1}{\sum_{j}\sum_{p}\mathbb{1}[\mathbf{M_j}(p)=k]}\sum_{j}\sum_{p}\mathbb{1}[\mathbf{M_j}(p)=k]\, \mathbf{f}_j(p),
\label{eq:support_prototype}
\end{equation}
where \(\mathbf{f}_j(p)\in\mathbb{R}^C\) denotes the feature vector at pixel \(p\) of \(F(\mathbf{X}_j)\), and \(\mathbb{1}[\cdot]\) is the indicator function. We denote by $\hat\mu_k$ the $\ell_2$-normalized version of $\mu_k$.

To reduce noise in novel-class predictions, we assess how well each predicted region (\ie, a connected component of pixels sharing the same predicted class label in the segmentation mask) aligns with the corresponding class prototype: regions whose feature representations are more consistent with the support examples are less likely to be spurious. For a region \(r\) in a synthetic image whose initial predicted class is \(k\) (where \(k\) is a novel class), we first compute the mean feature of all pixels within that region and then normalize it:
\begin{equation}
\mathbf{\hat v}(r)=
\frac{\frac{1}{|\mathcal{R}(r)|}\sum_{q\in\mathcal{R}(r)} \mathbf{f}(q)}
{\left\|\frac{1}{|\mathcal{R}(r)|}\sum_{q\in\mathcal{R}(r)} \mathbf{f}(q)\right\|},
\label{eq:pixel_normalize}
\end{equation}
where \(\mathcal{R}(r)\) denotes the set of pixels belonging to region \(r\). We then compute the cosine similarity between this region representation and the corresponding class prototype:
\begin{equation}
s(r)=\mathbf{\hat v}(r)^\top \mathbf{\hat\mu_k}.
\label{eq:pixel_proto_sim}
\end{equation}

If \(s(r)\ge\lambda\), the region is considered consistent with the novel class prototype, and we keep its novel class prediction; otherwise, pixels in the region are marked as free (label value \(255\)) to be relabeled later. 
The mask $M_1$ after APF therefore contains trustworthy novel-class pseudo-labels and free pixels (value \(255\)).

\subsubsection{Adaptive Pseudo-label Relabeling}
\label{sec:ai}

After APF, the mask contains reliable novel class pseudo labels and a set of free regions (\eg, with value \(255\)). These free pixels typically correspond to mispredicted or low-confidence regions and should be adaptively relabeled using information from the support set and the image itself. To relabel these free regions, we begin by constructing a global prototype dictionary 
\(\mathcal{G}=\{(k,\mathbf{\hat\mu_k})\}\), where \(\mathbf{\hat\mu_k}\) are the normalized support prototypes 
introduced in \cref{sec:APF}. For each synthesized or real image \(x\), we instantiate a local prototype 
dictionary by copying the global one.

Using the feature representation of image \(x\), we then estimate image-local prototypes for every class \(k\) that appears:
\begin{equation}
\mathbf{\mu_k^{(x)}} = \frac{1}{\sum_{p}\mathbb{1}[\mathbf{M_1}(p)=k]}
\sum_{p}\mathbb{1}[\mathbf{M}_1(p)=k]\, \mathbf{f}(p),
\label{eq:image_proto}
\end{equation}

To account for both global knowledge and image-specific statistics, we fuse the global and image-local
prototypes using a blending parameter \(\beta\in[0,1]\). This results in an adaptive prototype for image \(x\):
\begin{equation}
\tilde\mu_k^{(x)} = 
\beta\,\hat\mu_k + (1-\beta)\,\hat\mu_k^{(x)} ,
\label{eq:adaptive_proto_update}
\end{equation}
where \(\beta\) controls the influence of the global prototype relative to the image-local one. We denote by $\hat\mu_k^{(x)}$ the $\ell_2$-normalized version of $\mu_k^{(x)}$.

If no image local prototype can be computed for a novel class (\eg, no retained pixels remain for that novel class),
the adaptive prototype naturally defaults to the global one. In contrast, for base classes, we rely solely on the image-local prototypes. For every free pixel \(p\), we evaluate its feature representation against all class prototypes in the local dictionary:
\begin{equation}
s_k(p)=\mathbf{v}(p)^\top \tilde\mu_k^{(x)},\qquad k\in\mathcal{C}
\label{eq:freepixel_sim}
\end{equation}
where $\mathcal{C}$ denotes the set of candidate classes (novel classes and optionally base classes).
For each pixel $p$, we assign it to the class with the highest similarity score
$k^\star(p)=\arg\max_{k\in\mathcal{C}} s_k(p)$ only if the maximum similarity
$s^\star(p)=\max_{k\in\mathcal{C}} s_k(p)$ exceeds a confidence threshold $\lambda$;
otherwise, the pixel remains unlabeled.

The resulting mask \(\mathbf{M_2}\) after APR therefore contains retained pseudo-labels, 
newly relabeled regions, and any remaining free regions.

\subsection{SAM-based Boundary Refinement}
\label{sec:SABR}
\begin{figure}[t]
    \centering
    \includegraphics[width=\columnwidth]{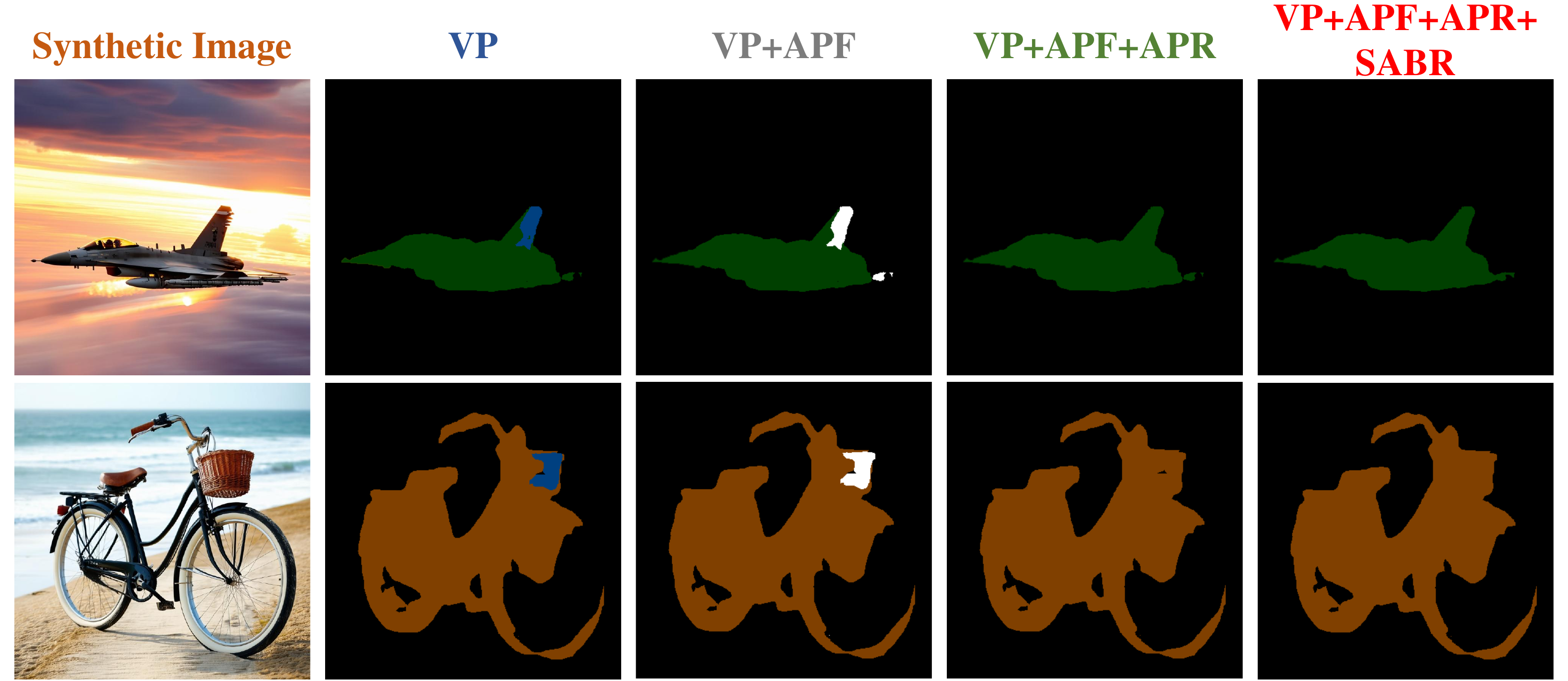}
    \caption{Quantitative ablation results showing the progressive improvements brought by the mask-quality refinement modules: APF, APR, and SABR.}
    \label{fig:component}
\end{figure}

Although $\mathbf{M_2}$ produced by APE preserves high-confidence pseudo-labels
and fills many free pixels, object boundaries are often imprecise and lack
fine-grained shape details, which may propagate noise into training and hinder
the learning of accurate spatial structures. To mitigate this issue, we apply
SABR to refine boundary localization and produce the final training mask
$\mathbf{M_3}$. Given the synthesized RGB image
$\mathbf{I}:\Omega\rightarrow\mathbb{R}^3$ with pixel domain
$\Omega\subset\mathbb{Z}^2$ and the intermediate mask
$\mathbf{M_2}:\Omega\rightarrow\{0,1,\dots,K\}\cup\{255\}$ (where $0$ denotes
background, $\{1,\dots,K\}$ semantic classes, and $255$ free pixels), we
initialize $\mathbf{M_3}=\mathbf{M_2}$ and process each class independently.
For a given class $k$, we first identify its boundary pixels as those labeled
as $k$ that are adjacent to pixels of a different label, compute a tight
bounding box enclosing these boundary pixels, and feed the original image
together with this box to SAM to obtain a binary foreground prediction
$S_k$ in the full-image coordinate frame. To prevent SAM from overwriting
high-confidence interior regions, we restrict updates to an uncertain region
defined as the union of a narrow boundary band and free pixels,
\begin{equation}
\mathcal{U}_k=\{p\in\Omega \mid \operatorname{dist}(p,\mathcal{B}_k)\le\delta
\ \text{or}\ \mathbf{M_2}(p)=255\}.
\end{equation}
where $\operatorname{dist}(\cdot,\cdot)$ denotes the minimum Euclidean distance
to the boundary set and $\delta$ controls the band width. To resolve overlaps
between classes, we process classes in descending order of pixel area in
$\mathbf{M_2}$ and update only pixels in $\mathcal{U}_k$ that have not been
assigned by higher-priority classes, assigning label $k$ to pixels predicted
as foreground by SAM and marking the rest as free. By constraining SAM to tight
candidate boxes and limiting updates to uncertain regions, $\mathbf{M_3}$
retains reliable interior labels while leveraging SAM to sharpen object
boundaries and fill free pixels, yielding pseudo-labels with substantially
improved boundary fidelity for subsequent segmentation training. As shown in \cref{fig:component}, APF and APR improve pseudo-label reliability,
while SABR refines object boundaries, jointly leading to progressive performance
gains.

%% file: sec/4_experiments.tex
\definecolor{mygreen}{RGB}{0,190,0}

\section{Experiments}

\begin{table*}[t]
\centering
\renewcommand{\tabcolsep}{3.mm}
\caption{Comparison with state-of-the-art methods on PASCAL-$5^i$ and COCO-$20^i$ under the GFSS setting. \textbf{Bold} and \underline{underlined} indicate the best and second best methods. Mean denotes the average of the mIoU over base and novel classes, while Harmonic represents their harmonic mean. \textbf{All methods use a ResNet-50 backbone and follow the same training/test split.}}
\label{tab:gfss}
\resizebox{\textwidth}{!}{%
\begin{tabular}{lcccccccc}
\toprule
\multirow{4}{*}{Methods} & \multicolumn{8}{c}{\textbf{PASCAL-$5^i$}} \\ \cmidrule(lr){2-9}
& \multicolumn{4}{c}{\textbf{1-shot}} & \multicolumn{4}{c}{\textbf{5-shot}} \\
\cmidrule(lr){2-5} \cmidrule(lr){6-9}
 & Base & Novel & Mean & Harmonic & Base & Novel & Mean & Harmonic \\
\midrule
CANet~\cite{zhang2019canet} (CVPR'2019) & 8.73 & 2.42 & 5.58 & 3.79 & 9.05 & 1.52 & 5.29 & 2.60 \\
PFENet~\cite{tian2020prior} (TPAMI'2020) & 8.32 & 2.67 & 5.50 & 4.04 & 8.83 & 1.89 & 5.36 & 3.11 \\
PANet~\cite{Wang2019PANetFI} (ICCV'2019) & 31.88 & 11.25 & 21.57 & 16.63 & 32.95 & 15.25 & 24.10 & 20.85 \\
SCL~\cite{Zhang2021SelfGuidedAC} (CVPR'2021) & 8.88 & 2.44 & 5.66 & 3.83 & 9.11 & 1.83 & 5.47 & 3.05 \\
MiB~\cite{Cermelli2020ModelingTB} (CVPR'2020) & 63.80 & 8.86 & 36.33 & 15.56 & 68.60 & 28.93 & 48.77 & 40.70 \\
CAPL~\cite{tian2022generalized} (CVPR'2022) & 64.80 & 17.46 & 41.13 & 27.51 & 65.43 & 24.43 & 44.93 & 35.58 \\
BAM~\cite{Lang2022LearningWN} (CVPR'2022) & 71.60 & 27.49 & 49.55 & 39.73 & 71.60 & 28.96 & 50.28 & 41.24 \\
DIaM~\cite{hajimiri2023strong} (CVPR'2023) & 66.79 & 27.36 & 47.08 & 38.82 & 64.05 & 34.56 & 49.31 & 44.90 \\
POP~\cite{Liu2023LearningOP} (CVPR'2023) & 46.68 & 19.94 & 33.32 & 27.94 & 41.50 & 36.26 & 38.80 & 38.70 \\
AlignDiff~\cite{qiu2024aligndiff} (ECCV'2024) & 67.32 & \underline{43.34} & 55.33 & \underline{52.73} & 68.43 & 47.41 & 57.92 & 56.01 \\
VP~\cite{hossain2024visual} (CVPR'2024) & \underline{74.58} & 34.99 & 54.79 & 47.63 & \textbf{74.86} & 50.34 & 62.60 & 60.20 \\
BCM~\cite{Sakai2024ASS} (NIPS'2024) & 71.15 & 41.24 & \underline{56.20} & 52.22 & 71.23 & \underline{55.36} & \underline{63.29} & \underline{62.30} \\
\midrule
\rowcolor{gray!40}
\textbf{Syn4Seg (ours)} & \textbf{74.80} & \textbf{46.13} & \textbf{60.47 {\footnotesize\textcolor{mygreen}{(+4.27)}}} & \textbf{57.07 {\footnotesize\textcolor{mygreen}{(+4.85)}}} & \underline{74.26} & \textbf{58.72} & \textbf{65.49 {\footnotesize\textcolor{mygreen}{(+2.20)}}} & \textbf{65.58 {\footnotesize\textcolor{mygreen}{(+3.28)}}} \\
\bottomrule
\end{tabular}}
\vspace{0.8em}
\renewcommand{\tabcolsep}{3.mm}
\resizebox{\textwidth}{!}{%
\begin{tabular}{lcccccccc}
\toprule
\multirow{4}{*}{Methods} & \multicolumn{8}{c}{\textbf{COCO-$20^i$}} \\ \cmidrule(lr){2-9}

 & \multicolumn{4}{c}{\textbf{1-shot}} & \multicolumn{4}{c}{\textbf{5-shot}} \\
\cmidrule(lr){2-5} \cmidrule(lr){6-9}
 & Base & Novel & Mean & Harmonic & Base & Novel & Mean & Harmonic \\
\midrule

CAPL~\cite{tian2022generalized} (CVPR'2022) & 43.21 & 7.21 & 25.21 & 12.36 & 43.71 & 11.00 & 27.36 & 17.58 \\
BAM~\cite{Lang2022LearningWN} (CVPR'2022) & 49.84 & 14.16 & 32.00 & 22.05 & 49.85 & 16.63 & 33.24 & 24.94 \\
DIaM~\cite{hajimiri2023strong} (CVPR'2023) & 42.69 & 15.32 & 29.00 & 22.55 & 38.47 & 20.87 & 29.67 & 27.06 \\
POP~\cite{Liu2023LearningOP} (CVPR'2023) & 30.38 & 9.63 & 20.00 & 14.62 & 24.53 & 16.19 & 20.36 & 19.51 \\
AlignDiff~\cite{qiu2024aligndiff} (ECCV'2024) & 41.72 & \textbf{22.44} & 32.08 & \underline{29.18} & 41.84 & 27.92 & 34.88 & 33.49 \\
VP~\cite{hossain2024visual} (CVPR'2024) & \underline{51.55} & 18.00 & \underline{34.78} & 26.68 & \underline{51.59} & 30.06 & \underline{40.83} & \underline{37.99} \\
BCM~\cite{Sakai2024ASS} (NIPS'2024) & 49.43 & 18.28 & 33.85 & 26.69 & 49.88 & \underline{30.60} & 40.24 & 37.93 \\
\midrule
\rowcolor{gray!40}
\textbf{Syn4Seg (ours)} & \textbf{53.50} & \underline{21.32} & \textbf{37.41 {\footnotesize\textcolor{mygreen}{(+2.63)}}} & \textbf{30.49 {\footnotesize\textcolor{mygreen}{(+1.31)}}} & \textbf{53.24} & \textbf{35.80} & \textbf{44.52 {\footnotesize\textcolor{mygreen}{(+3.69)}}} & \textbf{42.33 {\footnotesize\textcolor{mygreen}{(+4.34)}}} \\
\bottomrule
\end{tabular}}
\label{tab:gfss_all}
\end{table*}

\subsection{Experimental Setup}
\noindent\textbf{Datasets.}\quad
We conduct experiments on two standard GFSS benchmarks \cite{tian2022generalized}: PASCAL-$5^i$ and COCO-$20^i$. Following the evaluation protocol in \cite{hajimiri2023strong}, we assess our model on the entire test set for PASCAL-$5^i$ and on 10K test images for COCO-$20^i$. The reported results represent the average performance over four dataset splits.

\para{Data Pre-processing.}
For the GFSS model \cite{hossain2024visual}, we follow the same experimental setup as DIaM \cite{hajimiri2023strong}. During the base training phase, images containing both base and novel categories are retained, while the pixels of novel classes are relabeled as background. In the fine-tuning phase, only novel classes are labeled in the support set, with base classes treated as background. After incorporating synthetic data generated by the generative model, the GFSS model is employed to predict segmentation masks for the synthetic images and for the background regions in the original training set. Subsequently, APE and SABR are applied exclusively to the masks of the generated images to enhance the quality of novel class segmentation.
For the semantic segmentation model \cite{cheng2022masked}, we construct a mixed training dataset by combining the generated novel class images and their corresponding masks with the original base class images and their augmented masks in a balanced ratio. This mixed dataset is then used to train the Mask2Former mdoel \cite{cheng2022masked}.

\para{Evaluation Metrics.}
We evaluate model performance using two standard metrics: the mean intersection-over-union (mIoU) and the harmonic mean of base- and novel class mIoU (hIoU) \cite{Huang2023PrototypicalKL}. Following \cite{hajimiri2023strong}, the overall GFSS score is computed by averaging the mIoU of base and novel categories, while hIoU provides a more balanced measure by penalizing large discrepancies between the two. This evaluation strategy helps alleviate dataset bias in benchmarks such as PASCAL-$5^i$ and COCO-$20^i$, where the number of base classes is typically three times that of novel classes.

\para{Implementation Details.}
We conducted our experiments using SD3.5 in conjunction with SAM-ViT-H. For SD3.5, we set the classifier-free guidance scale to 5 and employed 40 denoising steps to generate images with a resolution of 768×768. Specifically, the GFSS model used in our experiments is VP \cite{hossain2024visual}. For VP \cite{hossain2024visual}, we follow the experimental setup in \cite{hossain2024visual}. We employ ResNet-50 as our backbone \cite{He2015DeepRL} to extract features. The feature dimension used for computing features and prototypes is set to 256. 
\begin{figure*}[t]
    \centering
    \includegraphics[width=\textwidth]{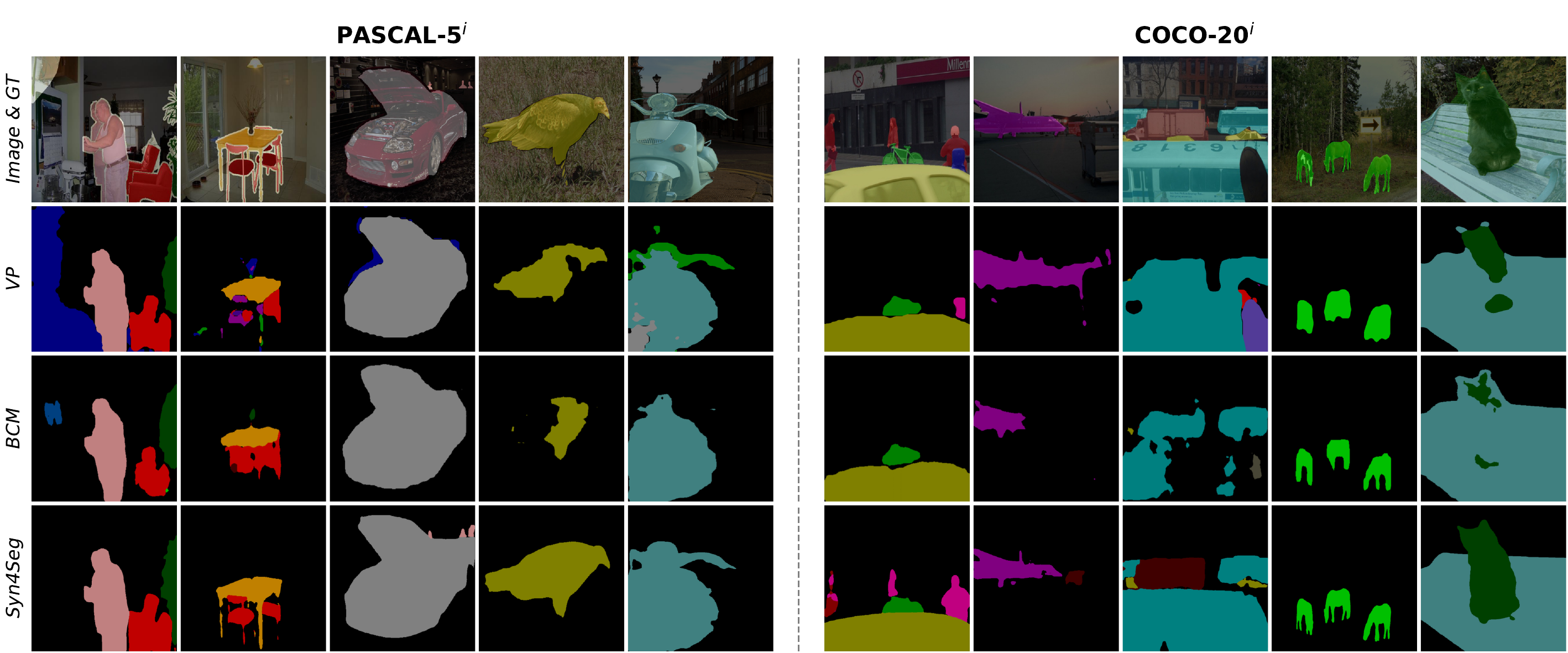}
    \caption{Qualitative results of the proposed Syn4Seg and state-of-the-art approach under the 1-shot setting. The left panel illustrates results on PASCAL-$5^i$, and the right panel on COCO-$20^i$. The first row shows the input images and their corresponding ground-truth masks, followed by the results of VP and BCM in the second and third rows, respectively. The fourth row presents the results of our Syn4Seg.}
    \label{fig:qualitative}
\end{figure*}

For the semantic segmentation model Mask2Former \cite{cheng2022masked}, we adopt the same experimental setup as \cite{cheng2022masked}. Specifically, ResNet-50 \cite{He2015DeepRL} is used as the backbone, and the model is trained using the per-pixel cross-entropy loss. The AdamW optimizer is used with a base learning rate of \(1 \times 10^{-4}\), a weight decay of 0.05, and a batch size of 16. The model is trained for 100 epochs on PASCAL-$5^i$ and 20 epochs on COCO-$20^i$. For all experiments, we empirically set the hyperparameters in \cref{sec:Syn4Seg_overview,sec:pas,sec:SABR} as follows: 
the threshold $\lambda=0.6$, the prototype blending coefficient $\beta=0.7$, respectively, and the band width for SABR as $\delta=5$. 
These values were chosen based on validation performance on the PASCAL-$5^i$ dataset 
and kept fixed for all experiments.

\subsection{Experimental Results}
\noindent\textbf{Comparison with State-of-the-arts.}\quad 
All experiments are conducted under the \emph{inductive} setting, where no unlabeled test data are used, as exploiting test data is unrealistic and harms generalization. \cref{tab:gfss_all} summarize the quantitative results on PASCAL-$5^i$ and COCO-$20^i$. Our model consistently and substantially surpasses existing methods: on PASCAL-$5^i$, it outperforms DIaM by 13.39\% mean mIoU and 18.25\% hIoU in the 1-shot setting, and by 16.18\% and 20.68\% in 5-shot. On COCO-$20^i$, the gains over DIaM reach 8.41\%/7.94\% (1-shot) and 14.85\%/15.27\% (5-shot). Compared with the strongest prior methods-BCM \cite{Sakai2024ASS} on PASCAL-$5^i$ and VP \cite{hossain2024visual} on COCO-$20^i$, our approach delivers further improvements of 4.27\%/4.85\% and 2.20\%/3.28\% on PASCAL-$5^i$, and 2.63\%/3.81\% and 3.69\%/4.34\% on COCO-$20^i$ under the 1-shot and 5-shot settings, respectively.

\para{Qualitative Results.}
\cref{fig:qualitative} compares the qualitative results of our method with VP~\cite{hossain2024visual} and BCM~\cite{Sakai2024ASS} on the PASCAL-$5^i$ and COCO-$20^i$ datasets under the 1-shot setting. 
Compared with VP and BCM, our method produces more coherent and complete segmentation masks, particularly for novel classes, while reducing fragmented predictions and spurious regions.

\subsection{Ablation Study}
\label{sec:ablation}
We conduct a comprehensive set of ablation studies on PASCAL-$5^i$ to analyze the contribution of each core component in Syn4Seg as well as the impact of key hyperparameters. Unless otherwise specified, all experiments are conducted under the 1-shot setting using ResNet-50 as the backbone.

\begin{table}[t]
\centering
\caption{Ablation study of different components in Syn4Seg, where rows (1)–(5) progressively add each component. \textbf{Bold} and \underline{underlined} indicate best and the second best method.}
\resizebox{0.9\columnwidth}{!}{
\begin{tabular}{ccccccccc}
\toprule
\textbf{ID} & \textbf{HDIG} & \textbf{APF} & \textbf{APR} & \textbf{SABR} & \textbf{Base} & \textbf{Novel} & \textbf{Mean} & \textbf{Harmonic} \\
\midrule
(1) &  &  &  &  &  73.68 & 41.75 & 57.22 & 52.48 \\
(2) & $\checkmark$ &  &  &  & \textbf{74.97} & 43.47 & 59.22 & 55.03 \\
(3) & $\checkmark$ & $\checkmark$ & -- & -- & 74.74 & 44.46 & 59.10 & 54.96 \\
(4) & $\checkmark$ & $\checkmark$ & $\checkmark$ & -- & 74.64 & \underline{45.65} & \underline{60.15} & \underline{56.65} \\
(5) & $\checkmark$ & $\checkmark$ & $\checkmark$ & $\checkmark$ & \underline{74.80} & \textbf{46.13} & \textbf{60.47} & \textbf{57.07} \\
\bottomrule
\end{tabular}
}
\label{tab:ablationofsyn4seg}
\vspace{-1em}
\end{table}

\begin{table}[t]
\centering
\caption{Ablation study on the influence of hyperparameters $\lambda$. \textbf{Bold} and \underline{underlined} indicate best and the second best method.}
\resizebox{0.6\columnwidth}{!}{
\begin{tabular}{ccccccccc}
\toprule
$\mathbf{\lambda}$ & \textbf{Base} & \textbf{Novel} & \textbf{Mean} & \textbf{Harmonic} \\
\midrule
 0.2 & 74.50 & 45.53 & 60.02  & 56.52 \\
 0.4 & 74.43 &\underline{45.66} & \underline{60.05} & \underline{56.60} \\
 0.6 & \textbf{74.80} & \textbf{46.13} & \textbf{60.47} & \textbf{57.07} \\
 0.8 & \underline{74.55} & 24.84 & 49.70 & 37.26 \\
 1.0 & 74.36 & 0.00 & 37.18 & 0.00 \\
\bottomrule
\end{tabular}
}
\label{tab:hyper}
\end{table}

\begin{table}[t]
\centering
\caption{Ablation study on the influence of hyperparameter $\beta$. \textbf{Bold} and \underline{underlined} indicate best and the second best method.}
\resizebox{0.6\columnwidth}{!}{
\begin{tabular}{ccccccccc}
\toprule
 $\beta$ & \textbf{Base} & \textbf{Novel} & \textbf{Mean} & \textbf{Harmonic} \\
\midrule
 0.1 & 73.05 & 45.31 & 59.18 & 55.93 \\
 0.3 & 74.47 & \underline{46.11} & 60.29 & \underline{56.95} \\
 0.5 & \textbf{74.95} & 45.79 & \underline{60.37} & 56.85 \\
 0.7 & \underline{74.80} & \textbf{46.13} & \textbf{60.47} & \textbf{57.07} \\
 0.9 & 74.64 & 45.53 & 60.09 & 56.56 \\
\bottomrule
\end{tabular}
}
\label{tab:beta}
\end{table}

\begin{table}[t]
\centering
\caption{Ablation study on the influence of hyperparameter $\delta$. \textbf{Bold} and \underline{underlined} indicate best and the second best method.}
\resizebox{0.6\columnwidth}{!}{
\begin{tabular}{ccccccccc}
\toprule
$\delta$ & \textbf{Base} & \textbf{Novel} & \textbf{Mean} & \textbf{Harmonic} \\
\midrule
 1 & 74.70 & 45.83 & 60.27 & 56.81 \\
 3 & \underline{75.14} & 45.73 & \underline{60.44} & 56.86 \\
 5 & 74.80 & \textbf{46.13} & \textbf{60.47} & \textbf{57.07} \\
 7 & \textbf{75.18} & 45.65 & 60.42 & 56.81 \\
 9 & 74.49 & \underline{46.11} & 60.30 & \underline{56.96} \\
\bottomrule
\end{tabular}
}
\label{tab:delta}
\end{table}

\begin{table}[t]
\centering
\caption{Ablation study on the influence of the backbone of Mask2Former. The \textbf{best} and \underline{second-best} results for each setting are highlighted.}
\resizebox{0.6\columnwidth}{!}{
\begin{tabular}{ccccccccc}
\toprule
 \textbf{Backbone} & \textbf{Base} & \textbf{Novel} & \textbf{Mean} & \textbf{Harmonic} \\
\midrule
 ResNet50 & \underline{74.80} & \underline{46.13} & \underline{60.47} & \underline{57.07} \\
 ResNet101 & \textbf{75.41} & \textbf{46.83} & \textbf{61.12} & \textbf{57.78} \\
\bottomrule
\end{tabular}
}
\label{tab:50vs101}
\end{table}

\begin{figure*}[t]
    \centering
    \setlength{\tabcolsep}{2pt}
    \renewcommand{\arraystretch}{1.0}
    \begin{tabular}{cccc}
        \includegraphics[width=0.235\textwidth]{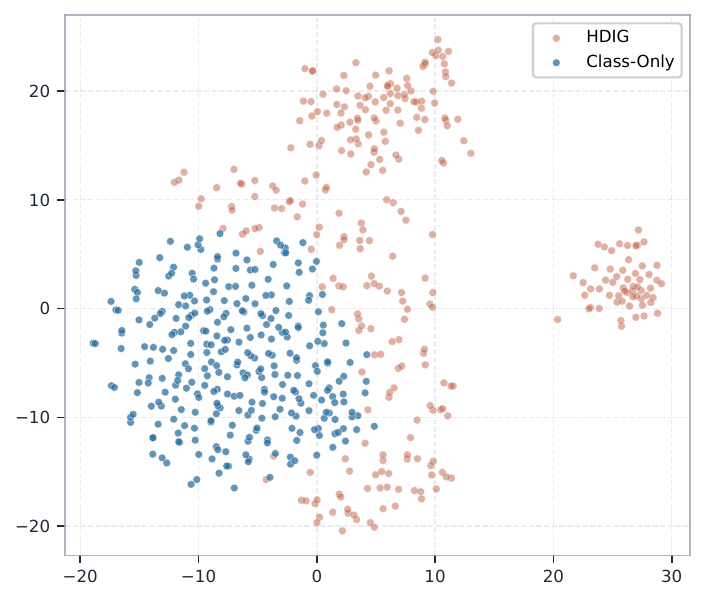} &
        \includegraphics[width=0.235\textwidth]{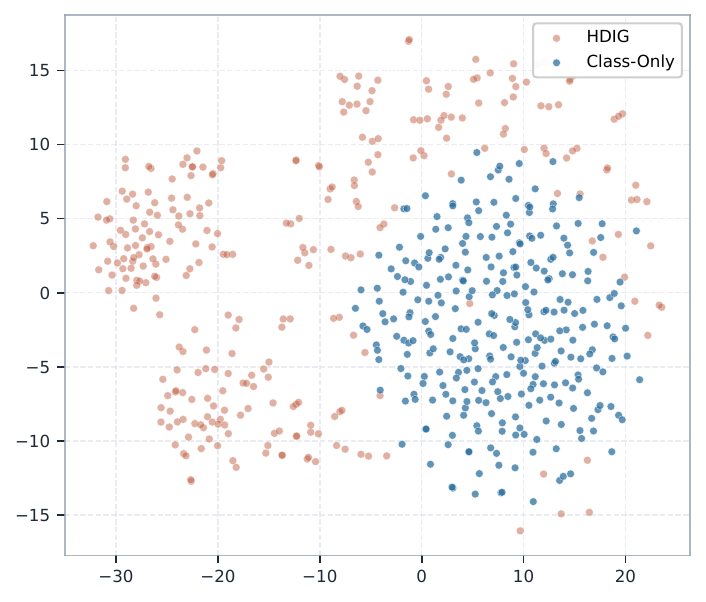} &
        \includegraphics[width=0.235\textwidth]{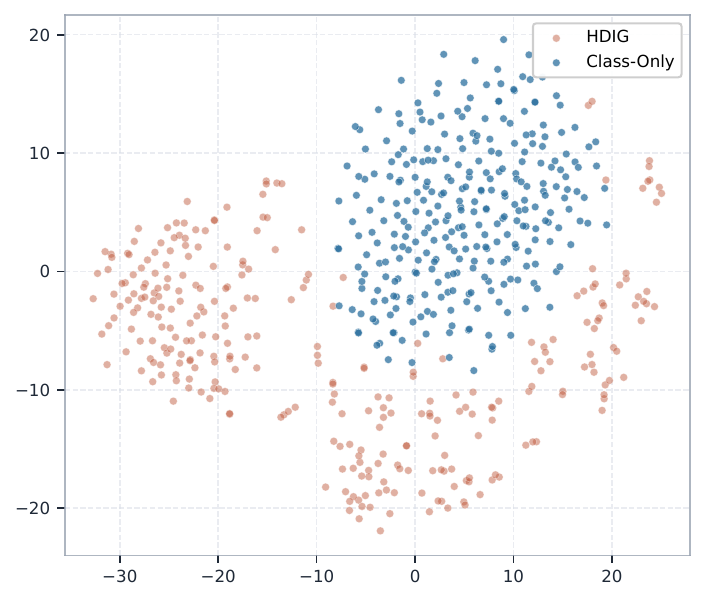} &
        \includegraphics[width=0.235\textwidth]{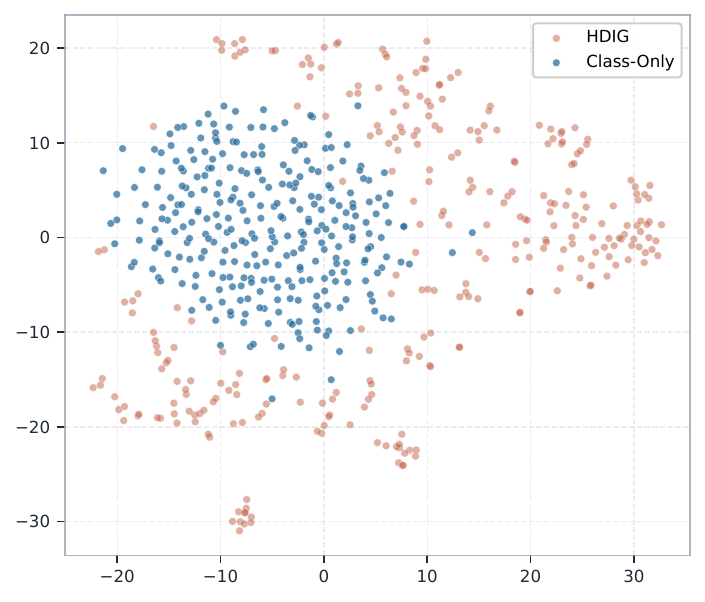} \\
        {\small bicycle} & {\small cow} & {\small horse} & {\small motorcycle} \\[2pt]
        \includegraphics[width=0.235\textwidth]{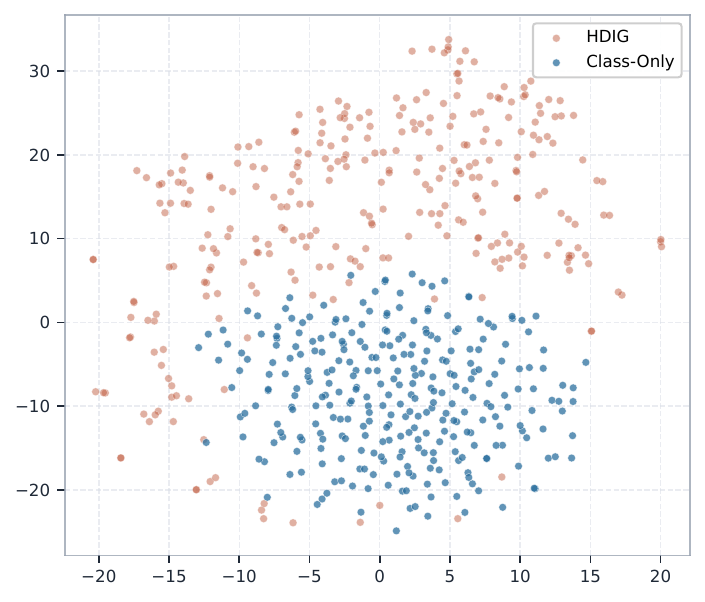} &
        \includegraphics[width=0.235\textwidth]{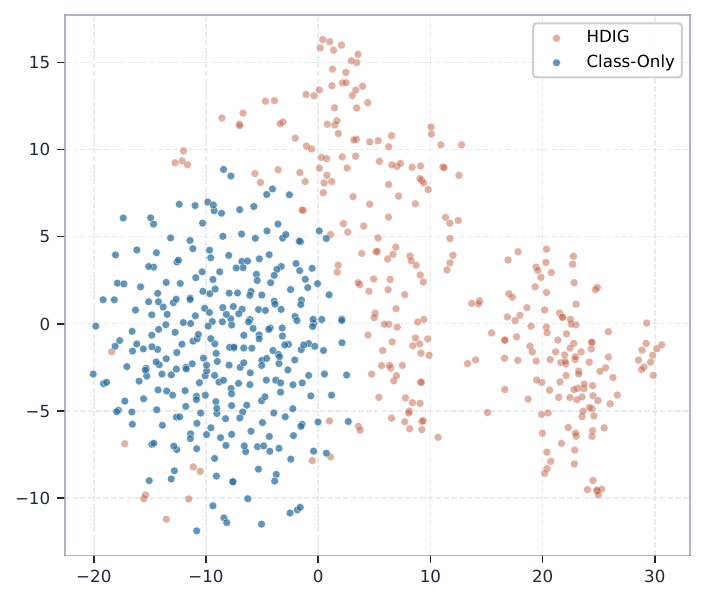} &
        \includegraphics[width=0.235\textwidth]{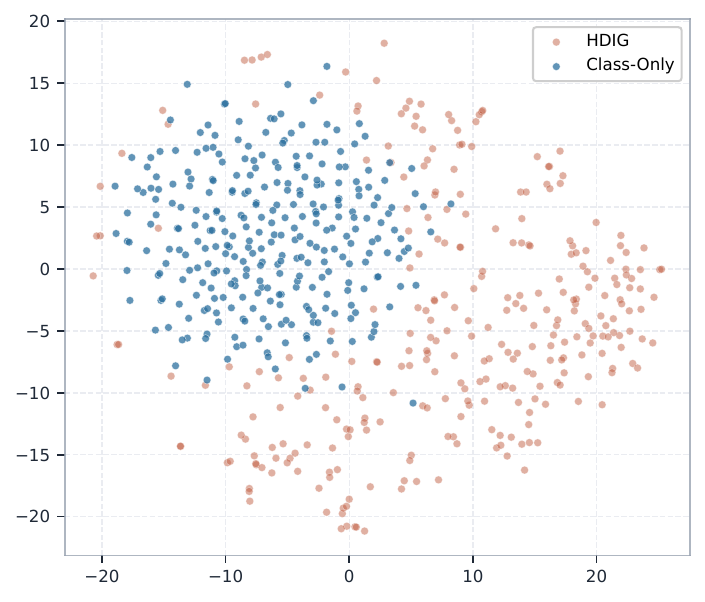} &
        \includegraphics[width=0.235\textwidth]{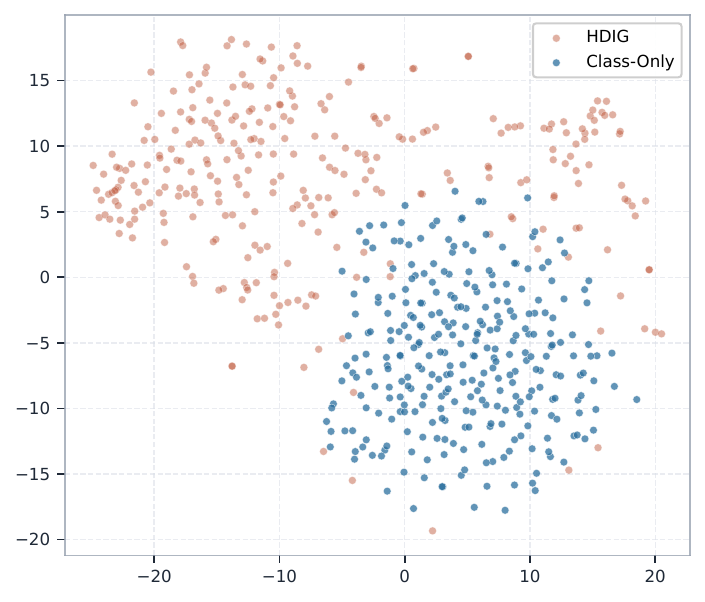} \\
        {\small pottedplant} & {\small sheep.pdf} & {\small sofa} & {\small tv}
    \end{tabular}
    \caption{t-SNE visualization of pooled ResNet-50 features extracted from images generated by class-only prompts and HDIG for eight categories.}
    \label{fig:tsne_hdig}
\end{figure*}

\para{Importance of Image Diversity.}
Comparing (1) and (2) in \cref{tab:ablationofsyn4seg}, introducing HDIG improves the diversity of generated novel images and yields a clear performance gain, raising novel mIoU from 41.75\% to 43.47\%. Meanwhile, the mean mIoU and harmonic mIoU are also improved. This verifies that enriching the visual diversity of synthesized novel samples provides broader appearance cues for downstream training, thereby improving generalization to novel classes. As further illustrated by the t-SNE visualization in \cref{fig:tsne_hdig}, features extracted from HDIG-generated images exhibit broader and more diverse distributions than those obtained from class-only prompting, indicating improved coverage of intra-class appearance variations.

\para{Effect of Mask Enhancement.}
As shown in \cref{tab:ablationofsyn4seg}, the APE module (APF+APR) improves mask quality by exploiting support information in a progressive manner. Specifically, APF suppresses misaligned regions in the pseudo masks, improving novel mIoU from 43.47\% to 44.46\%. Building on this, APR further relabels these ambiguous regions through refined correspondences, boosting novel mIoU to 45.65\% and improving both mean and harmonic mIoU. These results indicate that APE effectively enhances mask precision and semantic consistency through support-guided filtering and relabeling.

\para{Boundary Refinement with SABR.}
Incorporating SABR achieves the best overall performance, reaching 60.47\% mean mIoU and 57.07\% harmonic mIoU as shown in \cref{tab:ablationofsyn4seg}. Compared with the variant without SABR, the results show that explicitly modeling boundary-aware spatial reasoning is beneficial for resolving local ambiguities around object contours. This suggests that SABR can effectively refine coarse mask boundaries and improve segmentation consistency across both base and novel classes.

\para{Influence of the Hyperparameter $\lambda$.}
As shown in \cref{tab:hyper}, increasing $\lambda$ from 0.2 to 0.6 consistently improves novel mIoU, peaking at 46.13\%, while also yielding the best mean and harmonic mIoU. A single shared $\lambda$ is used throughout the framework, which keeps the design simple and avoids separate tuning for different components. This indicates that an appropriate $\lambda$ helps balance the contribution of synthesized information and improves novel-class generalization. However, when $\lambda$ is further increased, the performance drops sharply, reaching zero at $\lambda=1.0$, which suggests that overly large values may over-suppress useful information and severely damage generalization. Therefore, $\lambda=0.6$ provides the best trade-off in our setting.

\para{Influence of the Hyperparameter $\beta$.}
The effect of $\beta$ is reported in \cref{tab:beta}. As $\beta$ increases from 0.1 to 0.7, novel-class performance consistently improves, achieving the best novel mIoU of 46.13\% and harmonic mIoU of 57.07\% at $\beta=0.7$. This suggests that moderate feature interaction facilitates more effective cross-class knowledge transfer and enables the model to better exploit support information. However, when $\beta$ becomes larger, the performance shows a slight decline, indicating that excessive interaction may introduce less beneficial feature mixing. Overall, $\beta$ in the range of 0.5-0.7 yields stable and consistently strong results.

\para{Influence of the Hyperparameter $\delta$.}
As shown in \cref{tab:delta}, the performance remains relatively stable across a wide range of $\delta$, indicating that the proposed method is not highly sensitive to this parameter and exhibits good robustness to its variation. The best mean and harmonic mIoU are achieved at $\delta=5$, while $\delta=7$ yields the highest base mIoU, demonstrating a mild trade-off between base and novel performance. Therefore, considering the overall balance among different metrics, we set $\delta=5$ in all experiments.

\para{Influence of the Backbone.}
As shown in \cref{tab:50vs101}, replacing ResNet-50 with the deeper ResNet-101 backbone consistently improves performance across all metrics, indicating the general benefit of stronger feature extraction. In particular, novel mIoU increases from 46.13\% to 46.83\%, while mean mIoU and harmonic mIoU improve from 60.47\% to 61.12\% and from 57.07\% to 57.78\%, respectively. This suggests that a stronger backbone provides more discriminative feature representations and better captures fine-grained visual details, which is beneficial for downstream segmentation. At the same time, the performance gain remains moderate, indicating that our method does not heavily rely on backbone depth and still maintains strong effectiveness with a lighter architecture.

%% file: sec/5_conclu.tex
\section{Conclusion and Limitation}
In this paper, we present Syn4Seg, a generation-enhanced framework for generalized few-shot semantic segmentation that alleviates limited novel-class coverage under scarce annotations. By integrating diffusion-based image synthesis with support-guided pseudo-label enhancement and boundary-aware refinement, Syn4Seg generates high-quality, class-consistent image–mask pairs to strengthen novel-class supervision. Extensive experiments on PASCAL-$5^i$ and COCO-$20^i$ demonstrate consistent gains in both 1-shot and 5-shot settings while maintaining strong base-class performance, highlighting synthetic data as a scalable solution for GFSS.

\para{Future Work.}
Despite its promising performance, Syn4Seg has several limitations that warrant future investigation. First, the framework is inherently tied to the capabilities of the underlying generative and segmentation backbones, and may further benefit from continued progress in these foundation models. Second, although the prompt construction strategy promotes class-level diversity, it does not explicitly account for finer-grained factors, such as object pose, spatial configuration, or scene context. Third, the overall pipeline incurs additional offline computational cost due to large-scale synthesis and multi-stage pseudo-label refinement, motivating the development of more efficient generation and refinement schemes. Nevertheless, we believe that Syn4Seg provides a promising new perspective for GFSS, and that addressing these limitations offers valuable opportunities for future improvement.